\documentclass{egpubl}
\usepackage{pg2024}

\SpecialIssuePaper         

\CGFStandardLicense

\usepackage[T1]{fontenc}
\usepackage{dfadobe}  

\usepackage{cite}  
\BibtexOrBiblatex
\electronicVersion
\PrintedOrElectronic
\ifpdf \usepackage[pdftex]{graphicx} \pdfcompresslevel=9
\else \usepackage[dvips]{graphicx} \fi
\usepackage{url}
\usepackage{egweblnk}
\usepackage{amsmath}
\usepackage{graphicx}
\usepackage{booktabs}
\usepackage{bm}
\usepackage{multirow}
\usepackage{multicol}
\usepackage{color}
\usepackage{colortbl}
\definecolor{Red}{cmyk}{0,1,1,0}
\definecolor{Green}{cmyk}{1,0,1,0}
\definecolor{Cyan}{cmyk}{1,0,0,0}
\definecolor{Purple}{cmyk}{0.45,0.86,0,0}
\definecolor{Rosolic}{cmyk}{0.00,1.00,0.50,0}
\definecolor{Blue}{cmyk}{1.00,1.00,0.00,0}
\definecolor{Orange}{cmyk}{0,0.52,0.80,0}
\definecolor{Black}{cmyk}{1,0,0,1}

\newcommand{\best}{\cellcolor[rgb]{1,0.8,0.8}}
\newcommand{\second}{\cellcolor[rgb]{1,0.98,0.8}}

\title[]
      {Gaussian in the Dark: Real-Time View Synthesis From Inconsistent Dark Images Using Gaussian Splatting}

\author[S. Ye \& Z.H. Dong \& Y.B. Hu \& Y.H. Wen \& Y.J. Liu: Gaussian in the Dark]
{\parbox{\textwidth}{\centering 
        Sheng Ye$^{1}$$^*$\orcid{0000-0001-9280-6279},
        Zhen-Hui Dong$^{1}$$^*$\orcid{0009-0007-7879-8111},
        Yubin Hu$^{1}$\orcid{0000-0001-6107-2858},
        Yu-Hui Wen$^{2}$$^\dag$\orcid{0000-0001-6195-9782},
        Yong-Jin Liu$^{1}$\orcid{0000-0001-5774-1916}
        }
        \\
{\parbox{\textwidth}{\centering $^1$MOE-Key Laboratory of Pervasive Computing, the Department of Computer Science and Technology, Tsinghua University\\
         $^2$Beijing Key Laboratory of Traffic Data Analysis and Mining, School of Computer and Information Technology, Beijing Jiaotong University
       }
}
}

\begin{document}
\teaser{
 \vspace{-0.8cm}
 \includegraphics[width=0.85\linewidth]{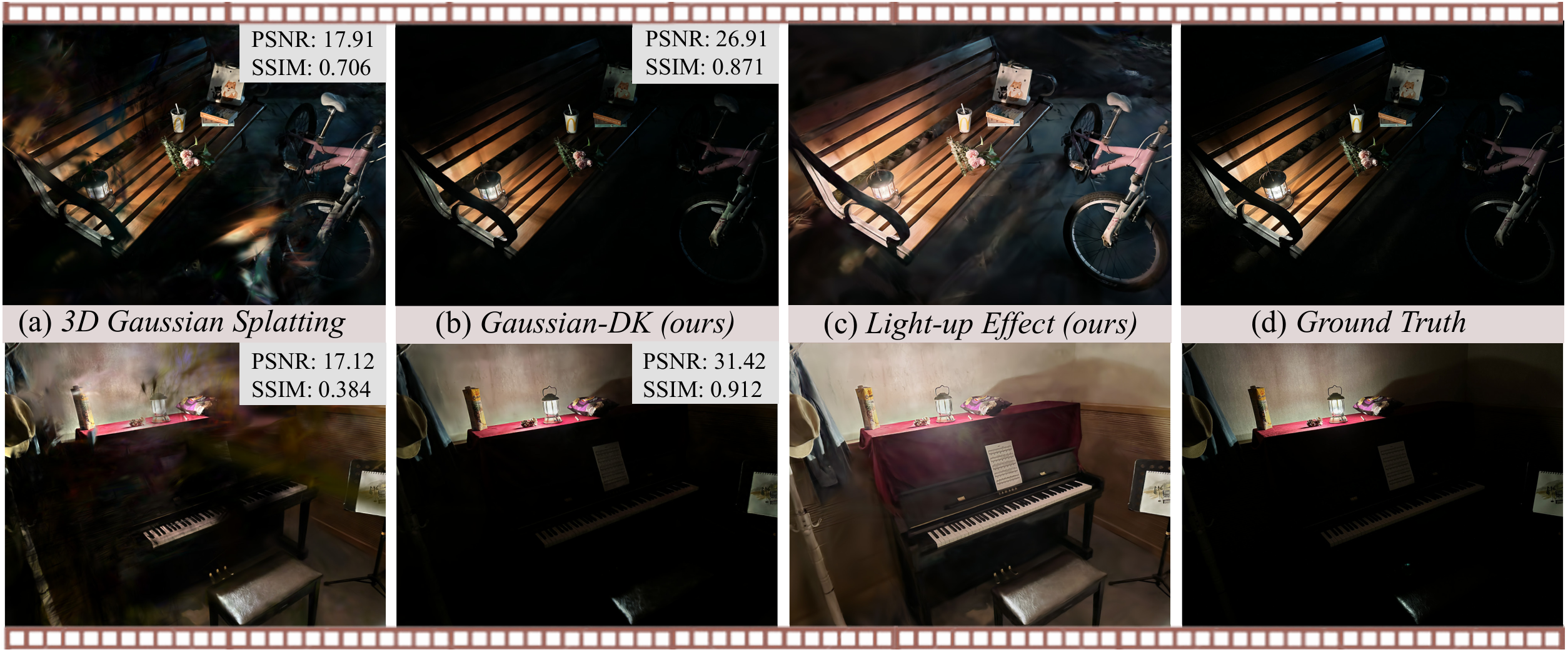}
 \centering
 \caption{Our proposed Gaussian-DK can produce high-quality renderings of novel views and synthesize exquisite light-up images from a set of inconsistent input photos captured in dark environments.}
 \label{fig:teaser}
}

\maketitle

\begin{abstract}
   3D Gaussian Splatting has recently emerged as a powerful representation that can synthesize remarkable novel views using consistent multi-view images as input. However, we notice that images captured in dark environments where the scenes are not fully illuminated can exhibit considerable brightness variations and multi-view inconsistency, which poses great challenges to 3D Gaussian Splatting and severely degrades its performance. To tackle this problem, we propose Gaussian-DK. Observing that inconsistencies are mainly caused by camera imaging, we represent a consistent radiance field of the physical world using a set of anisotropic 3D Gaussians, and design a camera response module to compensate for multi-view inconsistencies. We also introduce a step-based gradient scaling strategy to constrain Gaussians near the camera, which turn out to be floaters, from splitting and cloning. Experiments on our proposed benchmark dataset demonstrate that Gaussian-DK produces high-quality renderings without ghosting and floater artifacts and significantly outperforms existing methods. Furthermore, we can also synthesize light-up images by controlling exposure levels that clearly show details in shadow areas.


\end{abstract} 

\def\thefootnote{*}\footnotetext{These authors contributed equally to this work.}

\def\thefootnote{\dag}\footnotetext{Corresponding author}

\section{Introduction}
Novel view synthesis has always been an important task in computer vision and graphics. Existing methods \cite{nerf, mip-nerf, mip-nerf360, instant-ngp, 3dgs} can produce high-quality novel renderings using \textit{consistent} multi-view images captured under well-lit conditions. However, achieving such ideal lighting conditions is far from guaranteed in real-world scenarios, such as nighttime autonomous driving \cite{auto-driving} and navigation \cite{navi-dark}, low-light photography \cite{night-photography}, and robotic exploration in poorly illuminated environments \cite{auto-exploration}.

\begin{figure}[t]
  \centering 
  \includegraphics[width=\linewidth]{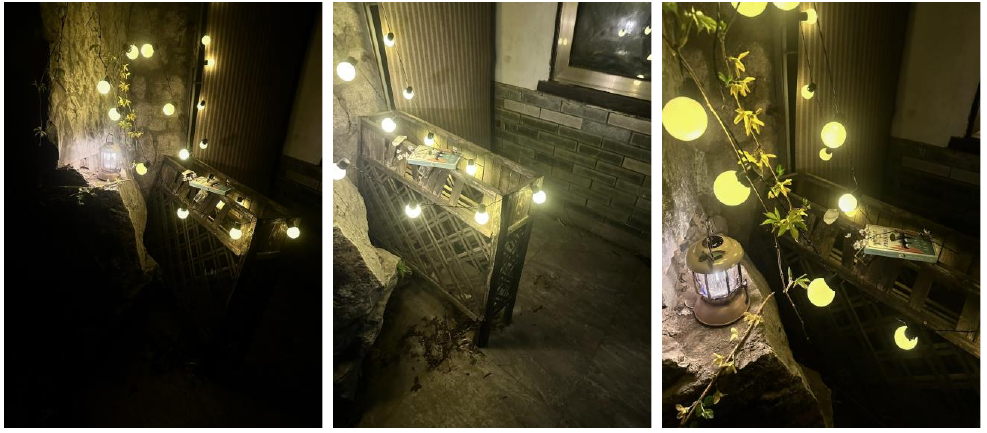}
  \caption{%
    The muti-view brightness inconsistencies of images captured in the dark environment.
  }
  \label{fig:inconsistency}
\end{figure}

We notice that brightness inconsistency generally appears when capturing images in a dark environment. Here, ``dark environment'' refers to a scene that is not thoroughly illuminated due to its complexity or vastness, rather than a totally black scene without any available light source. The light source in the dark scene is constant and static.
Such a dark environment is characterized by apparent brightness in near-light regions, but noticeable dimness or shadows in areas farther away or obstructed from the light.
Capturing images in dark environments is challenging because camera sensors can only process a limited range of luminance simultaneously \cite{dhr-cnn}, which makes it impossible to obtain an image that contains sufficient details in both bright and dark regions.
Thus, to acquire the full range of details, we have to constantly vary shooting angles and distances, while adjusting the focus positions across different viewpoints. 
This causes the camera to correspondingly adapt exposure levels to ensure clarity of the entire scene or focused subjects.
Specifically, when focusing on brighter areas, the camera automatically reduces the light intake to preserve highlight details, but also results in darker areas to be underexposed.
Conversely, when focusing on darker regions, the camera increases the light intake to capture details hidden in the shadows, but also causes overexposure in brighter regions. In conclusion, due to camera imaging characteristics, photos captured in dark environments exhibit brightness inconsistencies across different views (shown in Figure \ref{fig:inconsistency}).

3D Gaussian Splatting (3DGS) \cite{3dgs} has shown remarkable novel view synthesis quality with real-time rendering speed.
Nevertheless, we observe that it only works effectively under a well-conditioned, multi-view consistent setting, where the scene is entirely and uniformly illuminated.
As shown by our experimental results, the performance of 3DGS degrades severely with large floaters and ghosting artifacts, when presented with inconsistent image captures in aforementioned dark environments.
These problems can hinder the direct application of 3DGS in nighttime real-world scenarios.

In this work, we propose \textit{Gaussian-DK}, the first 3DGS method that can handle multi-view inconsistency in night photography.
Firstly, since the inconsistency is due to camera imaging characteristics, we explicitly distinguish the reconstruction of the physical world and the camera's imaging process. We leverage 3DGS to represent the radiance field of the physical world and incorporate exposure time, ISO gain, and aperture to model the exposure level for modulating the physical radiance value. Subsequently, we employ a convolutional neural network (CNN)
tone-mapper, which maps the modulated physical radiance values into image pixel values to achieve correct brightness effects.
Secondly, we attach a learnable light feature to each Gaussian, enhancing our method to better handle highlights and shadows.
Thirdly, we introduce a step strategy to scale the gradients of Gaussians based on their distance to the camera and successfully suppress the floaters near the camera, which are caused by complex brightness variations.
To evaluate the effectiveness of \textit{Gaussian-DK}, we propose a new benchmark dataset consisting of 12 scenes captured in dark environments.
Experiments show that \textit{Gaussian-DK} significantly outperforms 3DGS and other baselines while still achieving real-time rendering speed.
In summary, we make the following contributions:
\begin{itemize}
    \item We introduce a new method for recovering consistent scenes from inconsistent images captured in the dark. Moreover, we can synthesize light-up images by controlling exposure levels, which clearly display details in shadow areas.

    \item We design a comprehensive camera response modeling pipeline, a novel Gaussian light feature to achieve refined light and shadow effects, and a step-based gradient scaling strategy to suppress floaters near the camera.

    \item We propose a challenging dataset collected in real-world dark environments, demonstrating the efficacy of our method.
\end{itemize}

We also publicly release our code and the nighttime benchmark dataset at: \href{https://github.com/yec22/Gaussian-DK}{https://github.com/yec22/Gaussian-DK}.

\begin{figure*}[t]
  \centering 
  \includegraphics[width=0.95\linewidth]{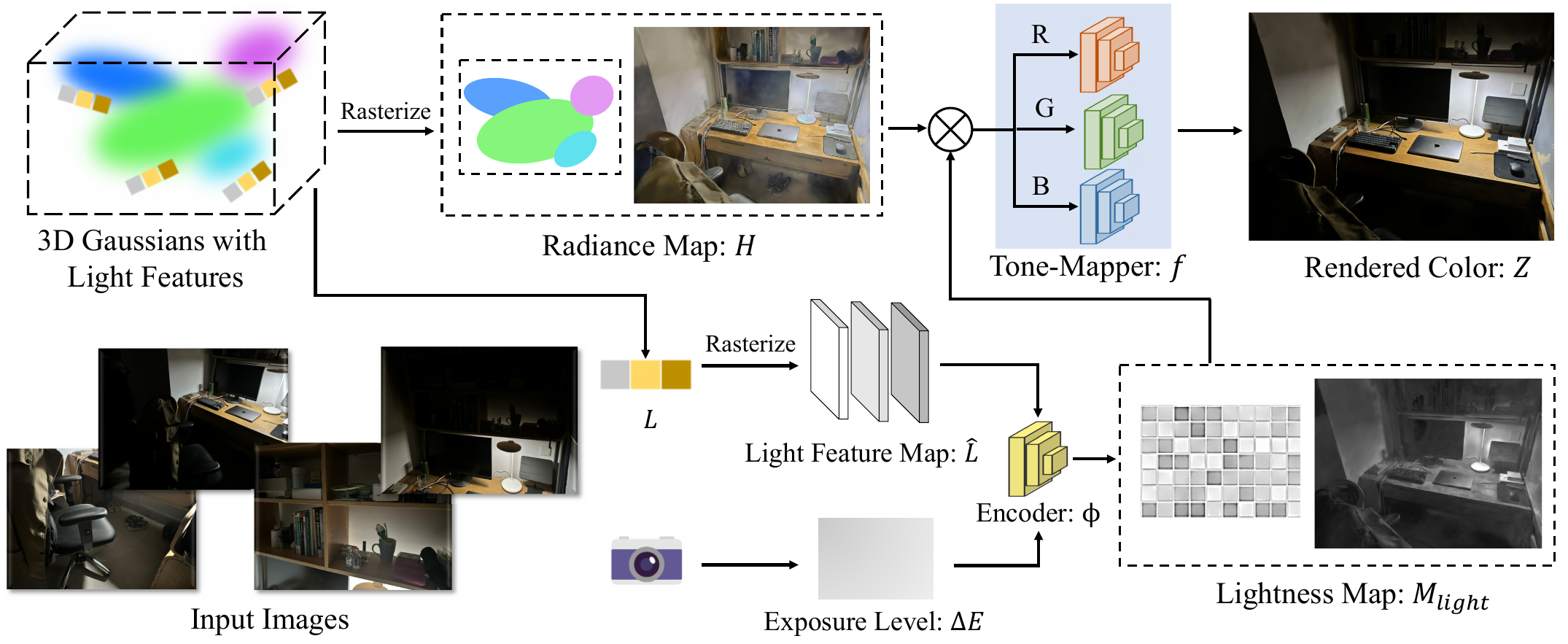}
  \caption{%
    The overall pipeline of \textit{Gaussian-DK}. We represent a consistent radiance field of the physical world using 3D Gaussians. Each Gaussian is additionally attached with a learnable light feature. Under a specified viewpoint, we rasterize the Gaussians to obtain a 2D light feature map and a radiance map. The feature map is combined with the exposure level to produce a pixel-wise lightness map, which is further used to modulate the rasterized 2D radiance map by pixel-wise multiplication. Finally, we utilize a CNN tone-mapper to convert the modulated radiance values into pixel values, achieving correct brightness effects.
  }
  \label{fig:pipeline}
\end{figure*}

\section{Related Works}
\subsection{Novel View Synthesis}
The novel view synthesis task aims to generate images of a specific subject or scene under unseen viewpoints. Traditional methods utilize light fields \cite{lightfield-4DCNN, light-field} or multi-view stereo \cite{mvsnet, r-mvsnet} to synthesize novel views.
Neural Radiance Field (NeRF) \cite{nerf} represents 3D scenes as implicit fields by mapping each spatial coordinate to density and color with a Multi-Layer Perceptron (MLP) network, and generates new views using volume rendering. Although NeRF notably outperforms previous methods, it requires dense images as input and a time-consuming training procedure. Various follow-up works attempt to further improve rendering quality \cite{mip-nerf, mip-nerf360, ref-nerf, zip-nerf}, increase training speed \cite{tensorf, DVGO, plenoxel, instant-ngp}, reduce the number of input views \cite{pixelnerf, ibrnet, diet-nerf, regnerf, depth-prior, freenerf}, and so on. However, NeRF and its variants still cannot achieve real-time rendering at high resolution due to the intensive sampling process in volume rendering.
The recently proposed 3D Gaussian Splatting \cite{3dgs} encodes the scene using numerous anisotropic 3D Gaussians and replaces the volume rendering with an efficient tile-based rasterizer, thereby getting rid of repetitive sampling and querying the neural network. As a result, 3DGS enables high-quality and real-time rendering at 1080p resolution. Nevertheless, the performance of 3DGS 
seriously degrades under inconsistent multi-view dark images, as shown by our experiments in \textit{Section} \ref{sec:quantitative}.

\subsection{Inconsistency Handling for Neural Rendering}
In real-world scenarios, the multi-view consistency assumption of captured images may be violated, making view synthesis more difficult.
NeRF-W \cite{nerfw} and some subsequent works \cite{neuralrecon-w, ha-nerf, cr-nerf, k-planes} aim to reconstruct different scenes from unconstrained internet photos, where the scene appearances are different.
To compensate for appearance variation and foreground occluders, NeRF-W incorporates an appearance embedding for each view and decomposes a scene into ``static'' and ``transient'' parts modeled by separate radiance fields.
VastGaussian \cite{vastgaussian} introduces a pixel-wise embedding and predicts a transformation map to adjust the rendered image, which handles the uneven illumination in large scenes.
The aforementioned works mainly cope with inconsistencies caused by changing light conditions or varying environments. In our problem setting, the environment and lighting stay still, and inconsistencies come from the camera imaging characteristics.
More relevant to our work, HDR-NeRF \cite{hdr-nerf} leverages a set of inconsistent images with different camera exposures to recover a consistent High Dynamic Range (HDR) radiance field. However, HDR-NeRF only considers the exposure time, while we model a comprehensive camera imaging process to compensate for more complex inconsistencies. Furthermore, we adopt 3DGS as a more powerful representation to achieve better rendering quality and real-time rendering speed.

\subsection{Dark Environment Settings}
Computer vision and graphics tasks in dark environments have attracted increasing attention recently, as dark environments are crucial for real-world applications and creating special visual effects. Yet, it also challenges the robustness and effectiveness of existing methods.
A few pioneering works have explored object detection \cite{yolo-dark}, SfM keypoint matching \cite{darkfeat}, robotic exploration \cite{darkgs}, image enhancement \cite{llnerf, aleth-nerf}, and video enhancement \cite{lv2018mbllenvideo, zhang2021video} in dark scenarios. In low-light computational photography, \cite{burstphotography} introduces a pipeline that captures, aligns,
and merges a burst of frames to reduce noise and increase dynamic
range. Subsequently, \cite{handheld-mobile} creates high-quality photographs in low light on a handheld mobile device.
In terms of view synthesis, RawNeRF \cite{rawnerf} first proposes to use multiple RAW-format images to generate novel views of nighttime scenes. However, RAW data requires high storage costs. Our \textit{Gaussian-DK} uses sRGB images as input for dark scene recovery, which could be useful when RAW data is not available.

\section{Method}
Clearly, a static dark environment is uniform and consistent in the physical world.
However, photos taken in such a consistent environment may still suffer from multi-view inconsistencies due to camera imaging characteristics.
Different camera exposure levels are employed across different views to ensure clarity of the photographed subjects, but also lead to significant image brightness variations that vanilla 3DGS cannot handle.

To address the above challenge, we propose \textit{Gaussian-DK} in this work. 
Our solution is to represent a consistent radiance field of the physical world using 3D Gaussians, and adopt a camera response module that maps radiance values to image pixel values while compensating for inconsistencies caused by camera imaging.
Furthermore, we design a gradient scaling strategy as an additional optimization regularizer to suppress floaters near the camera.
Figure \ref{fig:pipeline} shows the overall pipeline of our \textit{Gaussian-DK}.

\subsection{3D Gaussian Radiance Field}
We use a set of 3D Gaussians to represent the radiance field of the physical world. 
Each Gaussian can be parameterized by its mean position \(\mu\), covariance matrix \(\Sigma\), opacity \(\alpha\), and radiance \(c\) encoded by spherical harmonics.
Note that in vanilla 3DGS, \(c\) corresponds to pixel values ranging from 0 to 255, while in our \textit{Gaussian-DK}, \(c\) represents the physical radiance values ranging from 0 to +\(\infty\). By optimizing the learnable parameters and adaptively adjusting the density, these Gaussians can best represent the target scene. The 3D Gaussians are defined in world space as:
\begin{equation}
    G(x) = \exp \left( -\frac{1}{2} (x - \mu)^\top \Sigma^{-1} (x - \mu) \right).
\end{equation}
The rasterization process projects 3D Gaussians into a 2D radiance map. Given a specified camera pose, the 3D Gaussians within the camera frustum are first splatted into 2D by transforming the covariance matrix \(\Sigma\) to camera coordinates \(\Sigma'\). 
The value of the 2D radiance map \(H\) is then calculated by blending a set of \(N\) overlapping 2D Gaussians. The blending process can be written as:
\begin{equation}
\label{eqn:rasterize}
    H = \sum_{i \in N} c_i \alpha'_i \prod_{j=1}^{i-1} (1 - \alpha'_j), 
\end{equation}
where \(\alpha'_i\) is calculated as the multiplication of the opacity \(\alpha_i\) and the 2D projected Gaussian:
\begin{equation}
    \alpha'_i = \alpha_i \times \exp \left( -\frac{1}{2} (x' - \mu'_i)^\top {\Sigma'_i}^{-1} (x' - \mu'_i) \right),
\end{equation}
\(x'\) and \(\mu'_i\) are coordinates in the projected 2D space.

\subsection{Camera Response Modeling}
To prevent 3DGS from being affected by inconsistent input images and to recover a consistent scene, we design a camera response module that acts as a 3DGS modulate function to compensate for the inconsistency and achieve correct brightness effects. 
The camera response module consists of three components: exposure level condition (\textit{Section} \ref{EL}), learnable light feature refinement (\textit{Section} \ref{light-f}), and tone-mapping (\textit{Section} \ref{TM}).
Formally, our camera response module can be expressed as:
\begin{equation}
    Z = f(H, \Delta E, L), 
\end{equation}
where \(Z\) is the pixel value, \(H\) is the value of the 2D radiance map, \(\Delta E\) is the exposure level, which varies across different views, \(L\) is the light feature of Gaussian, and \(f\) is the tone-mapping function.

\subsubsection{Exposure Level Condition}
\label{EL}
We introduce the exposure level as a main condition to determine the overall brightness of the rasterized 2D radiance map of 3DGS. In photography, the exposure level refers to the total amount of light that enters the lens and reaches the sensor. Specifically, we model the exposure level using three camera parameters that can be easily obtained from the metadata (e.g., EXIF) of the captured image: exposure time, aperture, and ISO gain \cite{Digital-photography}. The aperture and exposure time control the amount of light that enters the camera, while the ISO gain, which is the sensitivity of the camera sensor to light, controls how much light the sensor utilizes.

During camera shooting, light flows in the form of photons towards the CMOS sensor and strikes the photodiode. The longer the exposure time, the more photons can enter the camera. Thus, the exposure level is proportional to the exposure time. The aperture can be regarded as the channel size of light travel, thereby the number of photons that the CMOS sensor receives is also proportional to the aperture area, determined by aperture diameter \(D\). In camera settings, the aperture value \(A\) is defined as \(A = \frac{v}{D}\), where \(v\) is the focal length.
Therefore, the exposure level is inversely proportional to the square of the aperture value. Under the same aperture and exposure time, doubling the ISO value will double the amount of light detected by the sensor, making the ISO value also directly proportional to the exposure level.
Combining all these relationships, we derive an empirical formula to calculate the exposure level \(\Delta E\):
 \begin{equation}
      \Delta E = \frac{K \times T \times ISO}{A^2},
 \end{equation}
where \(T\) is the exposure time, \(A\) is the aperture value, \(ISO\) is the sensitivity of the camera sensor and \(K\) is a constant. This formula incorporates exposure time, aperture, and ISO gain to reasonably represent exposure levels in digital photography.

\subsubsection{Gaussian Light Feature}
\label{light-f}
The above exposure level acts as a global brightness modifier during the camera imaging process.
Yet, we observe that using exposure level alone to adjust the overall brightness of an image is insufficient for modeling high-frequency lighting and shadow details in scenes with complex 3D geometries.
Specifically, when the exposure level is low, the rendered image becomes excessively dark, causing the shadow areas to lose details and the bright areas to appear blotchy. Conversely, when the exposure level is high, the highlight and shadow regions of the rendered image become slightly washed out, lacking clear texture and contrast.
This phenomenon indicates that different 3D locations should respond differently to the same exposure level.

To model the different corresponding response to the exposure level, we attach a learnable light feature vector for each Gaussian.
Under each viewpoint, we first rasterize the Gaussian light features to a 2D light feature map following Equation (\ref{eqn:rasterize}). Then, we adopt a CNN network to learn the response function between the 2D light feature and the exposure level, and consequently obtain a pixel-wise lightness map. This map precisely depicts the lightness level captured by the camera in the corresponding view. We use these lightness maps instead of exposure levels to modulate the physical radiance to compensate for different brightnesses. We formulate the lightness map \(M_{light}\) generation process as follows:
\begin{equation}
    M_{light} = \phi(\Delta E, \hat{L}),
\end{equation}
where \(\phi\) is the response function encoded by the CNN, \( \hat{L}\) is the rasterized 2D light feature map. By combining the light feature (for detail enhancement) with the exposure level (for high-level lighting adjustment), we can achieve fine-grained brightness control. This allows us to better represent light and shadow areas and preserve the rich texture of rendered images.

\subsubsection{Tone-Mapper}
\label{TM}
Finally, we modulate the rasterized 2D radiance map \(H\) of 3DGS by multiplying the above lightness map \(M_{light}\) to compensate for the pixel-wise brightness variations across different views. Subsequently, our tone-mapper tries to learn a function \(f\) that converts the modulated radiance values to image pixel values \(Z\):
\begin{equation}
      Z = f(H \times M_{light}).
\end{equation}
In practice, we notice that spherical harmonics are inherently low-frequency \cite{Sh-lighting, sh-stupid} and limited by a most applicable value range.
Directly using spherical harmonics to encode a complicated radiance field with drastic variations and unbounded values may lead to numerical instability during optimization. Thus, we represent all radiance values in the logarithmic domain. The logarithmic transformation compresses the value range and variation amplitude, making the optimization process of spherical harmonics much easier.
The final tone-mapping process can be formulated as:
\begin{equation}
    \ln(H) = \sum_{i \in N} c_i \alpha'_i \prod_{j=1}^{i-1} (1 - \alpha'_j),
\end{equation}
\begin{equation}
    Z = g\left(\ln(H) + \ln(M_{light})\right),
\end{equation}
where \(g = ( \ln f ^ { - 1 } ) ^ { - 1 }\) is the tone-mapping function in the logarithmic domain.
We utilize a lightweight channel-wise CNN to model the \(g\) function. Since the tone-mapping function of different color channels can be different, each channel (R, G, B) is processed separately using distinct network layers. Experimental results show that this channel-wise tone-mapper preserves the vividness and accuracy of textures in rendered images, and prevents color degradation compared to processing three color channels simultaneously using a global CNN. In addition, this lightweight CNN we adopt ensures the efficiency of the tone-mapping process, achieving good rendering quality without sacrificing too much rendering speed.

\subsection{Floater Removal}
Although introducing the camera response module can resolve most artifacts caused by inconsistencies, we note that a few floaters still appear near the camera. We assume that these floaters are Gaussians trying to compensate for complex brightness variations, which cannot be fully modeled by our camera response module. During training, these Gaussians are positioned as far away from the scene's center as possible to minimize their impact on other views, and thus typically appear closer to the camera. For scenes with drastic changes in capturing distances, these floaters can notably affect the final rendering results.

To suppress these floaters, we propose to decrease the gradients of Gaussians that are close to the camera, thereby preventing them from splitting or cloning during 3DGS's adaptive density control process. This is achieved by assigning a weight to the gradient magnitude of each Gaussian based on its distance to the camera calculated in Normalized Device Coordinates (NDC), inspired by ``Floaters No More''\cite{floater-no-more} and Pixel-GS \cite{pixel-gs}.
While the gradient scaling strategy is effective, we also observe that its suppression ability is too powerful, resulting in the loss of high-frequency details in views captured close to objects.
Therefore, we additionally incorporate the training iteration \(t\) to anneal the gradient scaling process, constraining strongly in the early stage to reduce floaters and annealing step by step to produce details.
Specifically, the step-based scaling weight \(w\) is defined as:
\begin{equation}
w(i, j) = 
\begin{cases} 
    \min\left(\left(\frac{z_{i,j}}{R}\right)^2, 1\right), & t < t_1 \\
    0.5 \times \min\left(\left(\frac{z_{i,j}}{R}\right)^2, 1\right), & t_1 \leq t < t_2 \\
    0, & t \geq t_2 
\end{cases}
\end{equation}
where \(z_{i,j}\) is the z-coordinate of Gaussian \(i\) in the NDC system under the \(j\)-th viewpoint, indicating the distance of this Gaussian to the camera, \(t_1\), \(t_2\) are the hyperparameters we set in advance and \(R\) represents the scale of the scene:
\begin{equation}
    R = 1.1 \times \max \left(\left\| {p}_i - \frac{1}{K} \sum_{j=1}^{K} {p}_j \right\|_2\right),
\end{equation}
where \(K\) is the number of training viewpoints of a scene, and \({p}_i\) represents the 3D position of the \(i\)-th camera center in the world coordinate system.

\begin{figure}[t]
  \centering 
  \includegraphics[width=\linewidth]{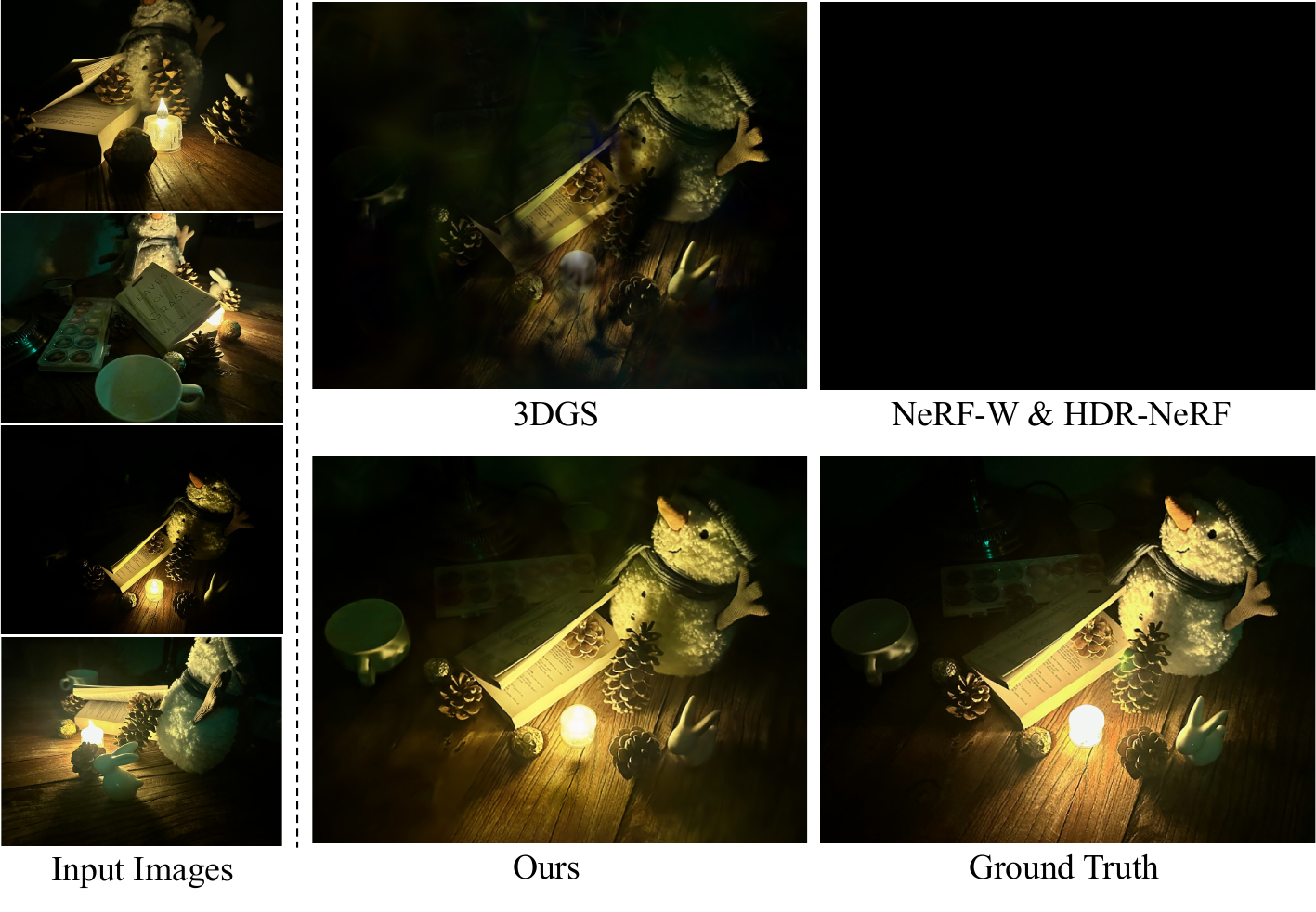}
  \caption{%
    The challenging ``Snowman'' scene with viewpoint variations and zoom-in capturing. While NeRF-based methods fail and 3DGS generates plenty of floaters, our method can still synthesize favorable images that are similar to Ground Truths.
   }
  \label{fig:snowman}
\end{figure}

\begin{table*}[tb]
\centering
\caption{Quantitative comparisons with baseline methods on our proposed dataset. We report three rendering quality metrics (PSNR, SSIM, LPIPS) and one rendering speed metric (FPS). The best and second-best are marked in red and yellow, respectively.}
\renewcommand\arraystretch{0.9}
\setlength{\tabcolsep}{1.5pt}
\begin{tabular}{c cccc cccc cccc cccc}
\toprule
\multirow{2.5}{*}{\textbf{Methods}} & \multicolumn{4}{c}{\textbf{Alley}} & \multicolumn{4}{c}{\textbf{Bicycle}} & \multicolumn{4}{c}{\textbf{Dormitory}} & \multicolumn{4}{c}{\textbf{Fence}} \\
\cline{2-17}
\rule{0pt}{10pt}
& PSNR$\uparrow$ & SSIM$\uparrow$ & LPIPS$\downarrow$ & FPS$\uparrow$ & PSNR$\uparrow$ & SSIM$\uparrow$ & LPIPS$\downarrow$ & FPS$\uparrow$ & PSNR$\uparrow$ & SSIM$\uparrow$ & LPIPS$\downarrow$ & FPS$\uparrow$ & PSNR$\uparrow$ & SSIM$\uparrow$ & LPIPS$\downarrow$ & FPS$\uparrow$ \\
\midrule
NeRF-W & \second{21.59} & \second{0.646} & 0.331 & 0.07 & \second{23.56} & \second{0.717} & 0.238 & 0.07 & \second{24.51} & \second{0.820} & \second{0.149} & 0.07 & \second{23.97} & \second{0.724} & \second{0.214} & 0.07 \\
HDR-NeRF & 17.87 & 0.558 & 0.488 & 0.05 & 19.95 & 0.608 & 0.267 & 0.05 & 21.23 & 0.684 & 0.279 & 0.05 & 19.66 & 0.578 & 0.372 & 0.05 \\
3DGS & 16.44 & 0.602 & \second{0.232} & \best{125} & 19.27 & 0.624 & \second{0.220} & \best{180} & 20.53 & 0.686 & 0.161 & \best{162} & 18.99 & 0.588 & 0.226 & \best{148} \\
Ours & \best{21.91} & \best{0.773} & \best{0.140} & \second{59} & \best{24.80} & \best{0.809} & \best{0.131} & \second{85} & \best{25.64} & \best{0.857} & \best{0.087} & \second{76} & \best{24.94} & \best{0.801} & \best{0.131} & \second{69} \\
\end{tabular}
\begin{tabular}{c cccc cccc cccc cccc}
\toprule
\multirow{2.5}{*}{\textbf{Methods}} & \multicolumn{4}{c}{\textbf{Flower}} & \multicolumn{4}{c}{\textbf{Kitchen}} & \multicolumn{4}{c}{\textbf{Livingroom}} & \multicolumn{4}{c}{\textbf{Piano}} \\
\cline{2-17}
\rule{0pt}{10pt}
& PSNR$\uparrow$ & SSIM$\uparrow$ & LPIPS$\downarrow$ & FPS$\uparrow$ & PSNR$\uparrow$ & SSIM$\uparrow$ & LPIPS$\downarrow$ & FPS$\uparrow$ & PSNR$\uparrow$ & SSIM$\uparrow$ & LPIPS$\downarrow$ & FPS$\uparrow$ & PSNR$\uparrow$ & SSIM$\uparrow$ & LPIPS$\downarrow$ & FPS$\uparrow$ \\
\midrule
NeRF-W & \best{21.66} & \second{0.662} & \second{0.223} & 0.07 & \second{24.84} & \second{0.812} & 0.205 & 0.07 & \second{22.43} & \second{0.749} & 0.243 & 0.07 & \second{25.44} & \second{0.803} & \second{0.157} & 0.07 \\
HDR-NeRF & 18.54 & 0.517 & 0.359 & 0.05 & 22.27 & 0.757 & 0.338 & 0.05 & 20.77 & 0.651 & 0.385 & 0.05 & 21.52 & 0.682 & 0.240 & 0.05 \\
3DGS & 16.17 & 0.541 & 0.226 & \best{133} & 18.68 & 0.699 & \second{0.196} & \best{181} & 19.97 & 0.728 & \second{0.167} & \best{142} & 18.78 & 0.527 & 0.248 & \best{214} \\
Ours & \second{21.50} & \best{0.728} & \best{0.135} & \second{64} &  \best{25.54} & \best{0.844} & \best{0.117} & \second{75} & \best{24.41} & \best{0.826} & \best{0.125} & \second{67} & \best{27.31} & \best{0.833} & \best{0.106} & \second{93} \\
\end{tabular}
\begin{tabular}{c cccc cccc cccc cccc}
\toprule
\multirow{2.5}{*}{\textbf{Methods}} & \multicolumn{4}{c}{\textbf{Street}} & \multicolumn{4}{c}{\textbf{Staircase}} & \multicolumn{4}{c}{\textbf{Snowman}} & \multicolumn{4}{c}{\textbf{Tree}} \\
\cline{2-17}
\rule{0pt}{10pt}
& PSNR$\uparrow$ & SSIM$\uparrow$ & LPIPS$\downarrow$ & FPS$\uparrow$ & PSNR$\uparrow$ & SSIM$\uparrow$ & LPIPS$\downarrow$ & FPS$\uparrow$ & PSNR$\uparrow$ & SSIM$\uparrow$ & LPIPS$\downarrow$ & FPS$\uparrow$ & PSNR$\uparrow$ & SSIM$\uparrow$ & LPIPS$\downarrow$ & FPS$\uparrow$ \\
\midrule
NeRF-W & \second{22.85} & \second{0.638} & 0.293 & 0.07 & \second{23.96} & \second{0.809} & \second{0.115} & 0.07 & -- & -- & -- & -- & \second{21.66} & \second{0.649} & 0.260 & 0.07 \\
HDR-NeRF & 21.20 & 0.577 & 0.424 & 0.05 & 18.07 & 0.566 & 0.343 & 0.05 & -- & -- & -- & -- & 18.58 & 0.504 & 0.397 & 0.05 \\
3DGS & 17.04 &  0.542 & \second{0.271} & \best{136} & 18.30 & 0.583 & 0.198 & \best{211} & \second{19.47} & \second{0.630} & \second{0.230} & \best{254} & 19.87 & 0.597 & \second{0.195} & \best{167} \\
Ours & \best{26.23} & \best{0.789} & \best{0.142} & \second{67} & \best{26.10} & \best{0.885} & \best{0.068} & \second{83} & \best{28.63} & \best{0.885} & \best{0.069} & \second{92} & \best{23.30} & \best{0.765} & \best{0.134} & \second{63} \\
\bottomrule
\end{tabular}
\label{tab:quantitative}
\end{table*}

\section{A New Benchmark Dataset}
For novel view synthesis in dark environments, little data is available. RawNeRF \cite{rawnerf} first introduces a denoising dataset that contains noisy RAW images captured at nighttime using a forward-facing setup (i.e., the disparities between viewpoints are relatively small).
Such a capture setup oversimplifies the problem and cannot meet the data collection needs in complicated real-world scenarios. Also, the RAW format results in huge storage requirements, limiting the number of images captured per scene.

To facilitate further research on novel view synthesis in dark conditions, we propose a new challenging dataset containing 12 real-world scenes (5 indoors and 7 outdoors). Each scene consists of approximately 80 to 130 sRGB images. Concretely, our indoor scenes include four room-level scenes (kitchen, living room, piano room, and dormitory) and one close-up scene (snowman). The outdoor scenes include two large-scale scenes (alley, street), three medium-scale scenes (bicycle, fence, and tree), and two close-up scenes (staircase, flower).

When collecting this dataset, we followed two main principles: 
1) The capture process should be as natural as possible. We did not manually adjust any camera parameters (exposure, ISO, white balance, etc.), avoiding artificial inconsistencies in captured photos. 
2) All details of the scene should be sufficiently recorded. Instead of adopting the forward-facing setup, we varied capture viewpoints significantly (including shooting angles and distances) to ensure that all subjects in dark scenes were clearly visible. Thus, our dataset contains plentiful wide-angle shots and close-up shots. 

We use an iPhone 14 Pro with auto-exposure mode for data collection. The image format is 8-bit JPEG with 3991 \(\times\) 2960 resolution. We downsample images by a factor of 4 for training.

\section{Experiments}
\subsection{Implementation Details}
We implement our \textit{Gaussian-DK} using the PyTorch framework and modify the rasterization CUDA kernels for our gradient scaling strategy.
We use COLMAP \cite{colmap} to estimate camera poses and to obtain the 3D positions of initial Gaussians.
Empirically, we set the annealing parameters \(t_1 = 4,000\) and \(t_2 = 7,000\). 
The dimension of each light feature vector is set to 8.
For the CNN that produces the lightness map and the CNN that tone maps each color channel, we employ three convolutional layers with \(hidden\_channel = 16\) and \(kernel\_size = 3\).
Following the adaptive density control of 3DGS, we densify and prune Gaussians every 100 iterations.
We increase one degree of the spherical harmonics every \(1,000\) iterations until it reaches three degrees.
For each scene, we optimize \textit{Gaussian-DK} for \(15,000\) iterations on a single NVIDIA A6000 GPU, which typically takes \(20\) minutes.

\begin{figure*}[htbp]
  \centering 
  \includegraphics[width=0.97\linewidth]{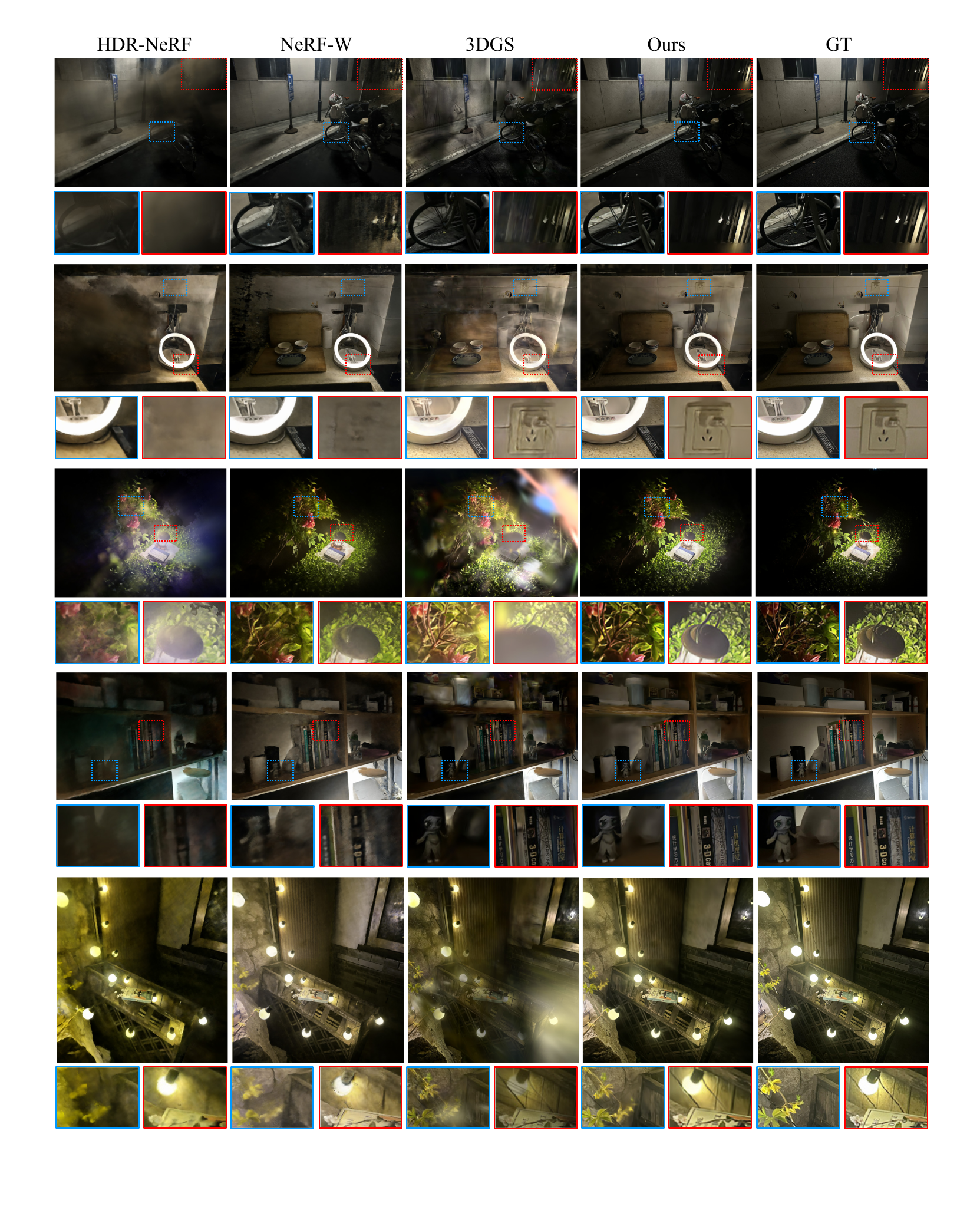}
  \caption{%
    Qualitative comparisons on our benchmark dataset. Our approach can eliminate floaters and produce images with rich details.
  }
  \label{fig:qualitative}
\end{figure*}

\subsection{Evaluation}
\subsubsection{Baselines and Metrics}
Our \textit{Gaussian-DK} first proposes to handle multi-view inconsistency in night photography, and thus, there is no feasible baseline.
To evaluate the effectiveness of \textit{Gaussian-DK}, we compare it with three most related methods, namely, vanilla 3DGS \cite{3dgs} and two NeRF-based methods for handling inconsistencies: NeRF-W \cite{nerfw} and HDR-NeRF \cite{hdr-nerf}.
For 3DGS and HDR-NeRF, we directly use their official code. Since the authors of NeRF-W do not release their code, we adopt a commonly used unofficial implementation$^1$.
\def\thefootnote{1}\footnotetext{\href{https://github.com/kwea123/nerf_pl/tree/nerfw/}{https://github.com/kwea123/nerf\_pl/tree/nerfw/}}
We exclude RawNeRF \cite{rawnerf} evaluation because it requires RAW-format images as input.

To evaluate the rendering quality, we use three quantitative metrics: Peak Signal-to-Noise Ratio (PSNR), Structural Similarity Index Measure (SSIM) \cite{ssim}, and Learned Perceptual Image Patch Similarity (LPIPS) \cite{lpips}.
We also report the rendering speed (Frames Per Second, FPS) of each method.

\begin{figure*}[t]
  \centering 
  \includegraphics[width=\linewidth]{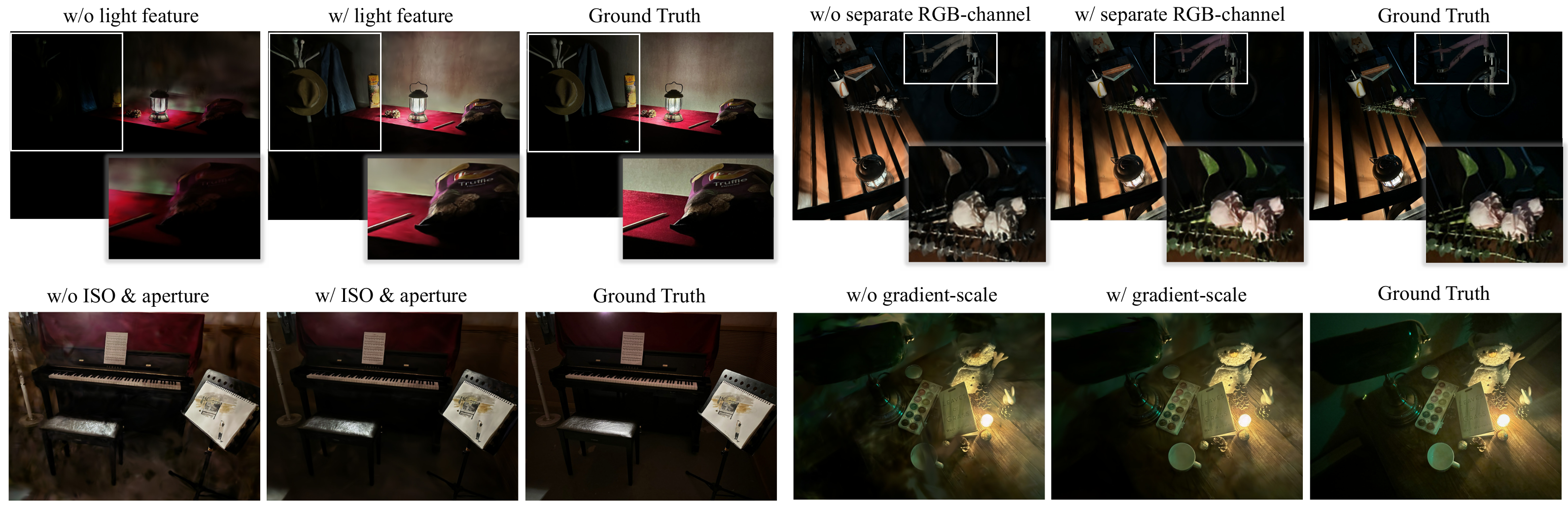}
  \caption{%
    Ablation results. Artifacts appear when any of our components are removed (recommend zooming in for better visualization).
  }
  \label{fig:ablation}
\end{figure*}

\subsubsection{Quantitative Comparison}
\label{sec:quantitative}
Quantitative results are summarized in Table \ref{tab:quantitative}.  \textit{Gaussian-DK} outperforms other baselines on PSNR, SSIM, and LPIPS in almost all scenes. Training the original 3DGS on inconsistent photo collections leads to poor results that are unable to compete with other methods for handling inconsistencies. In particular, \textit{Gaussian-DK} improves 3DGS by an average margin of 6.4dB on PSNR, 33\% improvements on SSIM, and 46\% on LPIPS.
NeRF-W achieves the second-best result, demonstrating that its per-view appearance compensation strategy can produce consistent renderings with a decent PSNR. However, its LPIPS and SSIM are significantly inferior to our method since NeRF-W lacks high-frequency textures and fine-grained details favored by these perceptual metrics. 
HDR-NeRF slightly outperforms 3DGS in terms of PSNR, indicating that this method can only handle limited input inconsistencies, as it only accounts for variations of exposure time.
Note that NeRF-based approaches output completely black renderings on the ``Snowman'' scene.
As shown in Figure \ref{fig:snowman}, we consider this close-up scene is too challenging, with large viewpoint variations and zoom-in capturing that make NeRF-based methods fail.
Additionally, NeRF-based methods perform poorly on FPS compared with 3DGS-based methods. We greatly improve the rendering quality while still achieving real-time rendering, but relatively slower than 3DGS.

\begin{table}[t]
\centering
\caption{Ablation results. Our full setup achieves the best performance in terms of PSNR, SSIM, and LPIPS.}
\setlength{\tabcolsep}{7pt}
\begin{tabular}{c ccc}
\toprule
\textbf{Config} & PSNR$\uparrow$ & SSIM$\uparrow$ & LPIPS$\downarrow$ \\
\midrule
w/o ISO \& aperture & 21.29 & 0.708 & 0.163 \\
w/o gradient-scale & 24.76 & 0.812 & 0.116 \\
w/o light feature & 24.15 & 0.803 & 0.132 \\
w/o separate RGB-channel & 24.64 & 0.807 & 0.121 \\
full model & \textbf{25.03} & \textbf{0.816} & \textbf{0.115}\\
\bottomrule
\end{tabular}
\label{tab:ablation}
\end{table}

\subsubsection{Qualitative Comparison}
Figure \ref{fig:qualitative} shows the visual comparison results of different approaches.
3DGS produces unpleasant floaters appearing as messy black or white splats that severely obscure the scene. These floaters are Gaussians trying to compensate for the varying brightness of each view, but do not agree with the scene geometry.
Both NeRF-based methods fall short of details and generate blurry results.
NeRF-W can recover a consistent but overly smooth scene. Scratched black artifacts may also appear in the edge regions of its renderings.
Moreover, when evaluating NeRF-W, we need to input the left half of the Ground Truth image to optimize the corresponding appearance embedding so that the rendered image can get the same brightness as Ground Truth. This optimization process is time-consuming, and the learned appearance embedding has poor interpretability.
HDR-NeRF generates large floater artifacts and synthesizes incorrect colors that deviate from Ground Truth, indicating that it is inadequate to handle complex inconsistencies.
In contrast, our method produces clean and visually appealing renderings that possess fine-grained details. Please see the supplementary demo video for more details.

\begin{figure}[t]
  \centering 
  \includegraphics[width=\linewidth]{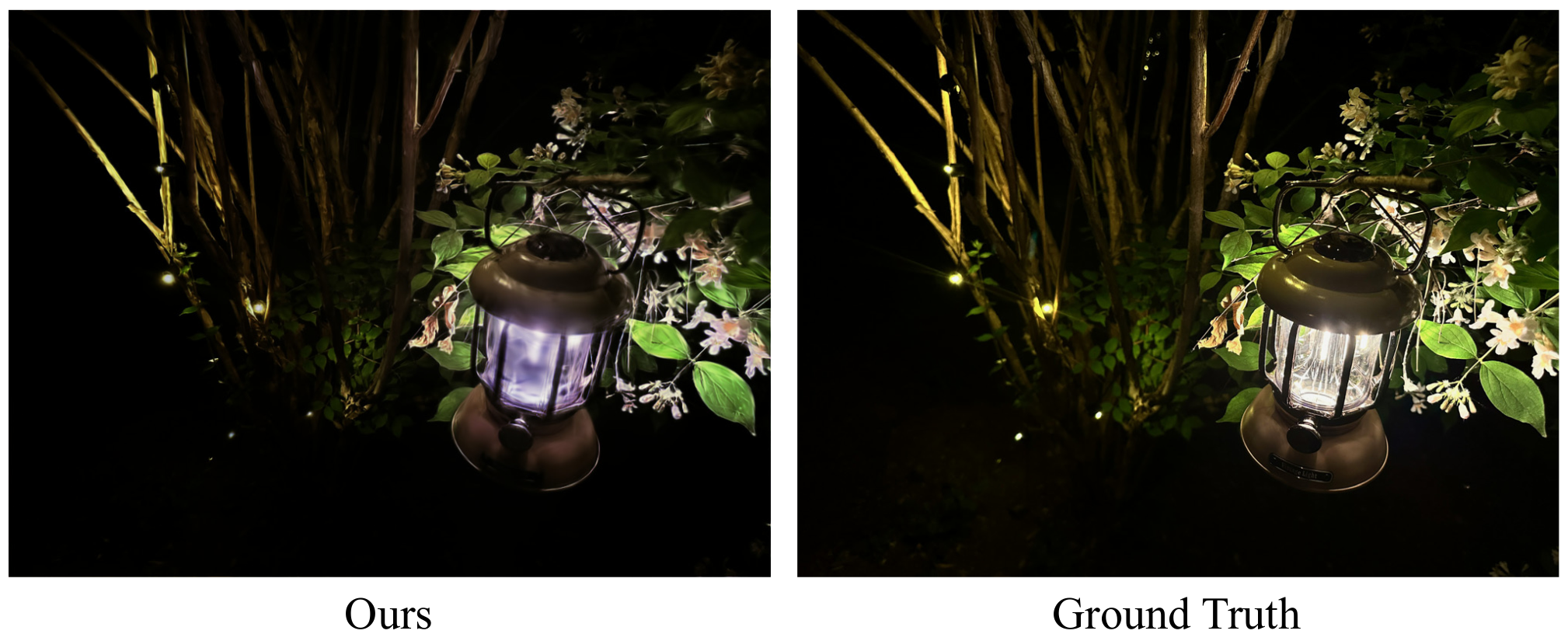}
  \caption{%
    Failure case of our approach.
   }
  \label{fig:failure}
\end{figure}

\begin{figure*}[t]
  \centering 
  \includegraphics[width=\linewidth]{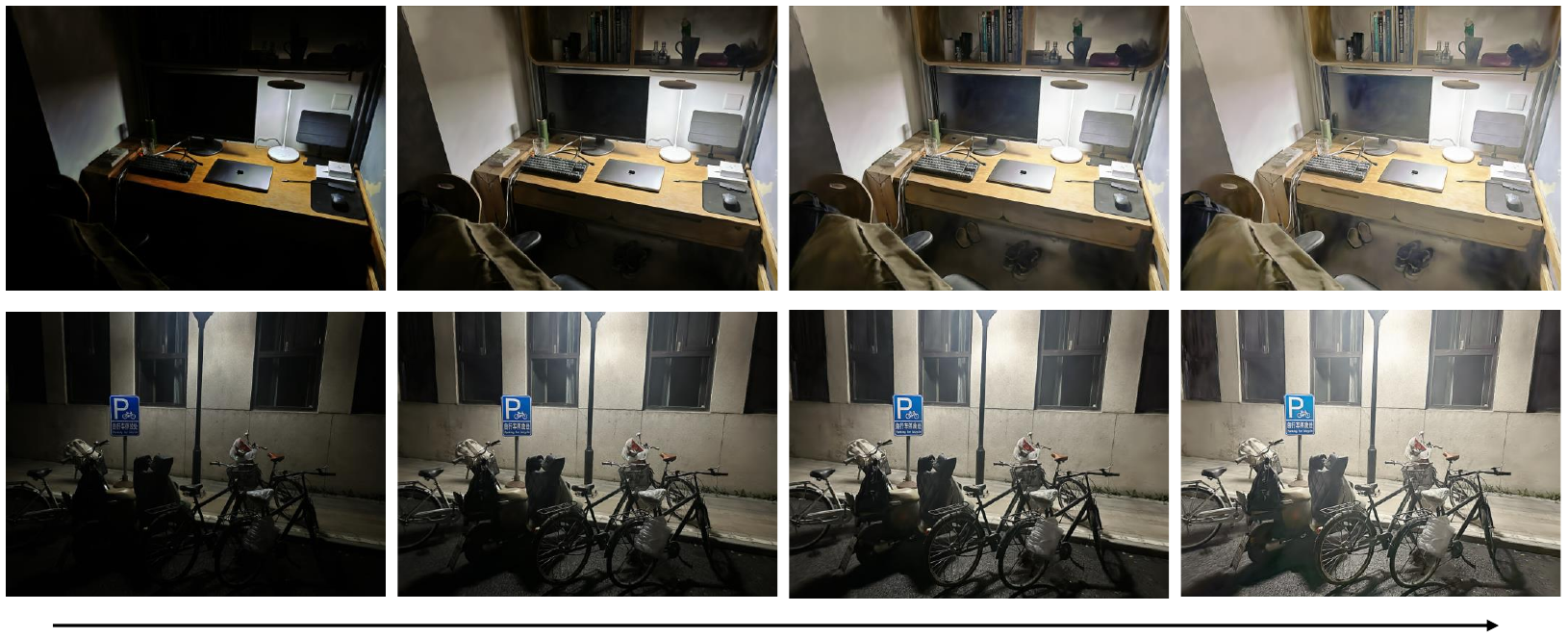}
  \caption{%
    We can synthesize images of different brightness by controlling the exposure level in our camera response module.
  }
  \label{fig:lightup}
\end{figure*}

\subsection{Ablations}
We evaluate our design choices by ablating four components of our method respectively: 1) light feature attached on each Gaussian (w/o light feature), 2) ISO gain and aperture considered in exposure level (w/o ISO \& aperture), 3) separate CNN used to tone map each color channel (w/o separate RGB-channel), 4) gradient scaling strategy based on camera distance (w/o gradient-scale).
Visual comparisons are shown in Figure \ref{fig:ablation}, and the averaged metrics are reported in Table \ref{tab:ablation}. 
Appending a light feature to each Gaussian can improve the details of objects hidden in shadows and delicately display highlight areas near the light source, thereby achieving more visually appealing effects.
Previous methods only consider exposure time in camera response modeling. Instead, we additionally consider ISO gain and aperture to obtain a more comprehensive camera response modeling. We have tried using only the exposure time to represent exposure level, which leads to considerably decreased performance with noticeable floater artifacts (shown in Figure \ref{fig:ablation}).
To prove the efficacy of our channel-wise CNN tone-mapper, we conduct an ablation that uses one global CNN to map RGB channels simultaneously. Although the quantitative results do not decrease too much, the visual effects are significantly inferior because renderings are washed out and lose vivid colors. When tone-mapping each color channel separately, we get lively renderings that are even better than the Ground Truths.
By removing the gradient-scaling strategy, we observe foggy renderings with small floaters appearing in areas close to the camera. With the gradient-scaling strategy, we can synthesize a clearer image.
Note that our full setup produces the best results, demonstrating the effectiveness of each component we proposed.

\subsection{Light-Up Visual Effect}
Figure \ref{fig:lightup} shows a bonus application of our approach.
In our camera response module, we adopt three camera parameters to model the exposure level (\textit{Section} \ref{EL}), which determines the overall brightness of rendered images.
By increasing the exposure level, we can light up the renderings to reveal more information hidden in the shadow regions, which is very useful in practice.

Note that in low-light areas, the supervision from sRGB images is weak and less informative, as shown in Figure \ref{fig:pipeline}. Therefore, Gaussians can be large, sparse, and low in transparency in low-light areas, as 3DGS does not need to generate dense Gaussians to represent these areas. For most dark images, these sparse Gaussians can produce decent renderings. When under some views with greater exposure levels, these less-dense Gaussians can lead to "cloudiness" artifacts. These artifacts can be further mitigated by adequate image capturing to ensure sufficient supervision.

\subsection{Limitation}
We note that our rendering quality will degrade when the exposure level of the rendered image significantly differs from those that appear in training images.
For example, the synthesized image shown in Figure \ref{fig:failure} has a much lower exposure level than all training views, and exhibits inaccurate brightness effects (darker than the Ground Truth, especially in light source regions).
We leave this problem as an interesting future direction.

\section{Conclusion}
In this work, we introduce \textit{Gaussian-DK}, a novel method to synthesize high-quality, consistent novel views from inconsistently dark input images while still achieving real-time rendering speed. Specifically, we exploit the powerful 3DGS to represent the radiance field of the physical world, and design a comprehensive camera response modeling pipeline. Moreover, we propose a gradient scaling strategy to control the splitting and cloning process of Gaussians, successfully suppressing floaters near the camera. Observing that multi-view data in dark environments is scarce, we collect a new challenging dataset containing 12 real-world scenes. Experiments on our dataset show that \textit{Gaussian-DK} notably surpasses existing baselines. Our approach can also synthesize light-up images that clearly reveal details in shadow areas.

\section*{Acknowledgments}
This work was supported by Beijing Science and Technology Plan Project (Z231100005923029) and The Talent Fund of Beijing Jiaotong University (2023XKRC045).

\bibliographystyle{eg-alpha-doi} 
\bibliography{egbibsample}

\end{document}